\documentclass[10pt,twocolumn,letterpaper]{article} 

\usepackage{avss}
\usepackage{times}
\usepackage{epsfig}
\usepackage{graphicx}
\usepackage{amsmath}
\usepackage{amssymb}

\usepackage{multirow,array}
\usepackage[table]{xcolor}
\usepackage{subfigure}
\usepackage[boxruled,vlined]{algorithm2e} 
\usepackage{dsfont}
\usepackage{lipsum}
\graphicspath{ {./figs/} }

\usepackage[pagebackref=false,breaklinks=true,letterpaper=true,colorlinks,bookmarks=false]{hyperref}
\usepackage{cite}

\avssfinalcopy 

\def\avssPaperID{0044} 

\ifavssfinal\fi
\begin{document}

\title{Active Collaborative Ensemble Tracking}

\author{Kourosh Meshgi, Maryam Sadat Mirzaei, Shigeyuki Oba, Shin Ishii\\
Graduate School of Informatics, Kyoto University\\
Yoshida-Honmachi, Sakyo Ward, Kyoto 606--8501, Japan\\
{\tt\small meshgi-k@sys.i.kyoto-u.ac.jp}
}

\maketitle

\begin{abstract}
A discriminative ensemble tracker employs multiple classifiers, each of which casts a vote on all of the obtained samples. The votes are then aggregated in an attempt to localize the target object. Such method relies on collective competence and the diversity of the ensemble to approach the target/non-target classification task from different views. However, by updating all of the ensemble using a shared set of samples and their final labels, such diversity is lost or reduced to the diversity provided by the underlying features or internal classifiers' dynamics. Additionally, the classifiers do not exchange information with each other while striving to serve the collective goal, i.e., better classification. In this study, we propose an active collaborative information exchange scheme for ensemble tracking. This, not only orchestrates different classifier towards a common goal but also provides an intelligent update mechanism to keep the diversity of classifiers and to mitigate the shortcomings of one with the others. The data exchange is optimized with regard to an ensemble uncertainty  utility function, and the ensemble is updated via co-training. The evaluations demonstrate promising results realized by the proposed algorithm for the real-world online tracking.
\end{abstract}

\section{Introduction}
\label{sec:intro}
Visual tracking is one of the fundamental problems in computer vision, having a broad range of applications from human-computer interfaces, to automatic surveillance, video description/editing/indexing, and autonomous navigation systems. Generative trackers attempt to construct a robust object appearance model, or to learn it on-the-fly using advanced machine learning techniques such as subspace learning \cite{ross2008incremental}, hash learning \cite{fang2017online}, dictionary learning \cite{taalimi2015online}, and sparse learning \cite{bao2012real}. On the other hand, discriminative models focus on target/background separation using correlation filters \cite{kiani2015correlation,danelljan2015learning,danelljan2016beyond} or dedicated classifiers \cite{nam2016learning}, which assist them to dominate the visual tracking benchmarks \cite{wu2013online}. Tracking-by-detection methods primarily treat tracking as a detection problem to avoid having model object dynamics especially in the case of sudden motion changes, extreme deformations, and occlusions \cite{tang2007co,bai2013randomized}. 

\begin{figure}
\includegraphics[width=1\linewidth]{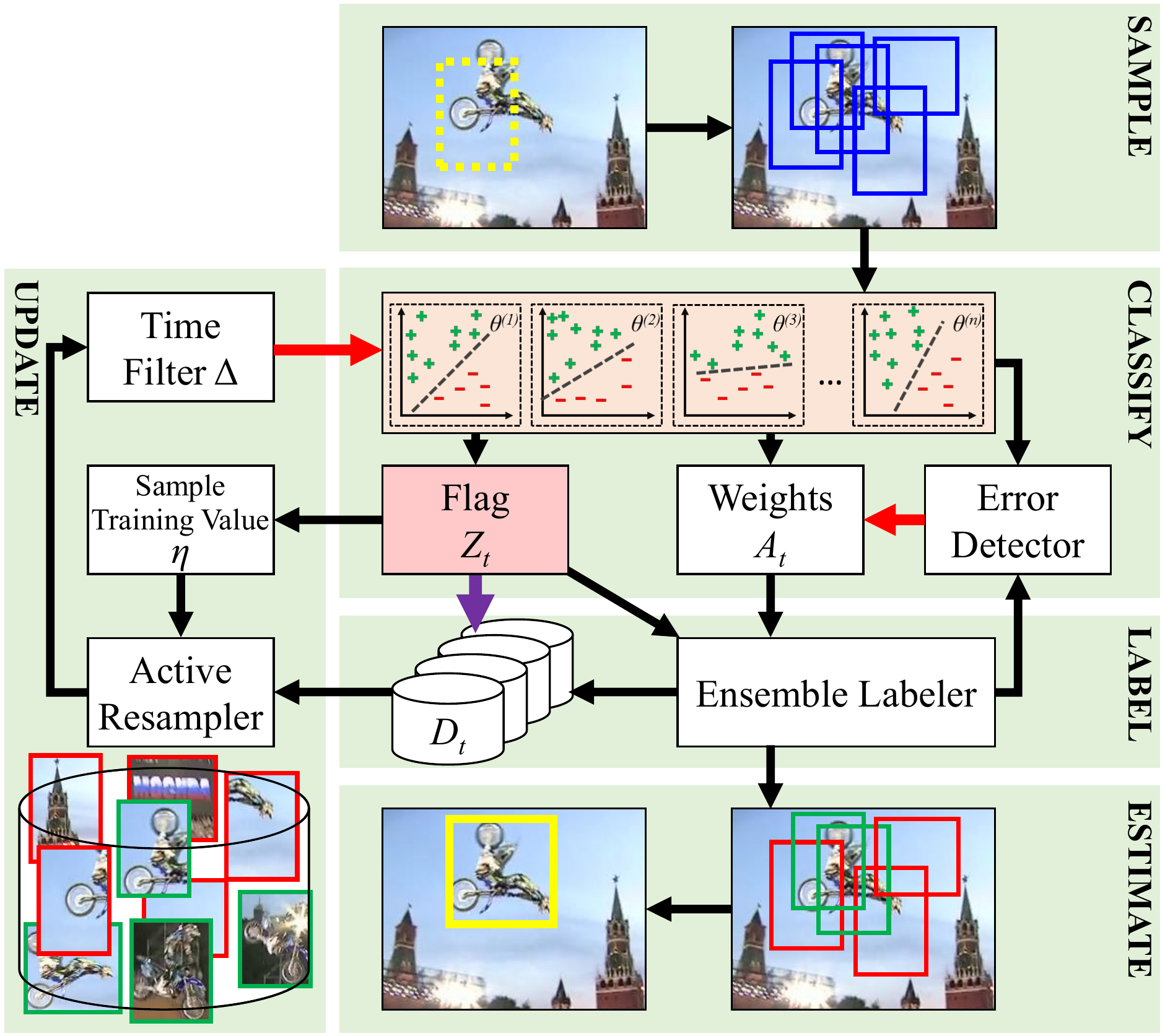}
\caption{Schematic of the proposed tracker, ACET. Black arrows indicate the flow of the information, red arrows represent update signals, purple arrow represent the co-learning procedure.}
\label{fig:schematic}
\vspace{-0.5 cm}
\end{figure}

However, these trackers are still vulnerable to illumination variation, in-plane and out-of-plane rotations, scale changes, and background clutter. Typical problems of tracking-by-detection schemes are \textit{(i)} label noise, where inaccurate labels confuses the classifier and and degrade the classification accuracy \cite{tang2007co}, \textit{(ii)} self-learning loop, in which the classifier are re-trained by their own output from earlier frames, thus accumulating error over time \cite{bai2013randomized}, \textit{(iii)} model drift, that is a side-effect of imperfect model update \cite{matthews2004template} and mismatch between model update frequency and target evolution rate \cite{grabner2008semi}, \textit{(iv)} equal weights for all samples in evaluating the target \cite{henriques2012exploiting} and training the classifier \cite{lapedriza2013all}, despite the uneven contextual information in different samples, and \textit{(v)} assuming stationary distribution of target, which does not hold for most of the real-world scenarios with drastic target appearance changes \cite{bai2013randomized}.

Ensemble tracking \cite{avidan2007ensemble,grabner2006real,oza2005online,saffari2010robust,babenko2009visual,saffari2009line,gall2011hough,leistner2010miforests,bai2012robust,zhang2014meem} and co-tracking \cite{tang2007co} frameworks provide effective frameworks to tackle one or more of these challenges. In such frameworks, the self-learning loop is broken, and the labeling process is performed by eliciting the belief of a group of classifiers (ensemble) or another classifier that has a stronger belief about the sample's label (collaborator). However, these frameworks typically do not address some of the fundamental demands of tracking-by-detection approaches like a proper model update to avoid model drift or non-stationary of the target sample distribution. Here, the non-stationarity means that the appearance of an object may change so significantly that a negative sample in the current frame looks more similar to a positive example in the previous frames. Besides, ensemble classifiers do not exchange information, and collaborative classifiers entirely trust the other classifier to label the challenging samples for them and are susceptible to label noise.

We propose an effective integration of ensemble tracking and co-tracking, which involves the merits of each while their complementary nature counteracts the demerits of each other. Here, an ensemble of trackers is employed to label a sample. Those classifiers that are uncertain about the label, are excluded from the final decision about the sample's label, and the rest of the classifiers perform a weighted voting for labeling the sample. The contributing classifiers are then retrained with the newly labeled samples, based on the concept of co-training. If the classifiers disagree each other for most of the samples, it is likely that the target is mostly occluded. The use of an ensemble would undermine label noise problem, while co-training breaks the self-learning loop, provides an effective model update, and enforce the diversity into the ensemble. By providing different memory spans for different members of the ensemble, the model update rate of the ensemble is automatically adjusted to the evolution rate of the target, and limited memory horizon resolves the issues with non-stationarity of the observations. By limiting the classifiers' retraining data to only the most informative ones (i.e., to assume different ``training values'' for samples), the non-stationarity is better addressed, the generalizability of the ensemble is improved, and speed of the tracker is boosted. 

We evaluated our proposed framework (active collaborative ensemble tracking or ACET) with other ensemble trackers and also the state-of-the-art in visual tracking on object tracking dataset \cite{wu2013online} to demonstrate the effectiveness of this method, and discussed its merits and demerits.

\section{Related Work}
\label{sec:bkg}

\textbf{Ensemble-based Tracking:} Using a (linear) combination of several (weak) classifiers with different associated weights has been proposed in a seminal work by Avidan \cite{avidan2007ensemble}. Align with this study, constructing an ensemble by boosting \cite{grabner2006real}, online boosting \cite{oza2005online,leistner2010miforests}, multi-class boosting \cite{saffari2010robust} and multi-instance boosting \cite{babenko2009visual,zhang2013real} provides better and better performance for ensemble trackers. The boosting may or may not couple with the ensemble changes such as feature adjustment \cite{gall2011hough} or addition/deletion of the ensemble's members \cite{grabner2006real,saffari2009line}. To date, boosting has been widely used in self-learning based tracking methods despite its low endurance against label noise \cite{santner2010prost}. An alternative way to tune the weights of the weights of an ensemble is via a Bayesian treatment \cite{bai2012robust}. Aside from using different features, the members of an ensemble may be constructed from randomized subsets of training data \cite{meshgi2016robust} or different time snapshots of a classifier evolving by time \cite{zhang2014meem}.

\textbf{Training Value of Samples:} Lapedriza et al. \cite{lapedriza2013all} discussed that different samples have different training value for a classifier, and using a wisely selected subset of samples for training/retraining the classifier outperforms the training with full dataset, for instance, due to mislabeled or inaccurately demarcated samples. Having a better training set for a tracking-by-detection classifier leads to enhanced generalization and faster convergence to the final performance which is suitable for converging to the piece-wise stationary target distribution (the distribution may change by every drastic change of target's appearance). To address this, researchers came by different approaches to provide good samples for tracking using context \cite{grabner2010tracking,kiani2017learning}, saliency maps \cite{kwon2017leveraging}, confidence maps \cite{tang2007co}, and optical flow \cite{kalal2012tracking}. Adaptive weights for the samples based on their appearance similarity to the target \cite{perez2002color}, occlusion state
\cite{meshgi2016data}, and spatial distance to previous target location \cite{wu2015robust}
have also been considered, however, selecting an efficient subset for classifier re-training have been ignored, as most of the trackers retrain on all of the data, a randomized subset of it \cite{meshgi2016robust}, or in special cases re-sample the training data based on their boosting value \cite{leistner2009robustness}. A ``clean'' subset of training samples to re-train the classifier can achieve much higher performance than the full set \cite{zhu2012we,razavi2012latent}, therefore, a principled ordering and selection of the samples reduces the cost of labeling and accelerate the performance with smaller re-training sample size \cite{vijayanarasimhan2011cost}. Different studies have tried to provide this small clean subset by different approaches: pruning outliers \cite{de2001robust} and hard-to-learn samples \cite{angelova2005pruning}, learning easy-to-classify examples first (as known as the Curriculum learning)\cite{bengio2009curriculum}, treating samples as noisy observations \cite{felzenszwalb2010object}, defining a training value for each sample by treating each sample as a separate classifier \cite{lapedriza2013all}, and robust loss functions for special classifiers (e.g., SVMs). Arguably, the most common setting is active learning, which selects the training samples to be labeled/selected at each step for higher gains in performance. Some approaches focus on learning the hardest examples first (e.g., those closest to the decision boundary), whereas some others gauge the information contained in the sample and select the most informative ones first. For instance, in the case of an ensemble of classifiers, the samples for which the ensemble disagrees the more, contains more information about how to train the ensemble. This concept is known as query-by-committee \cite{seung1992query} that tries to provide the best classifier with as few labeled instances as possible.

\section{Proposed Method}
\label{sec:method}

A tracking-by-detection algorithm usually estimates the target state $\mathbf{p}_t$ in time $t \in \{1,\ldots,T\}$ by obtaining several samples $\mathbf{p}^j_t \in \mathcal{P}_t$, scoring them $s^j_t \in \mathcal{S}_t$, labeling them $\ell^j_t \in \mathcal{L}_t$, and aggregating them, $\mathbf{p}_t = \psi (\mathcal{P}_t | \mathcal{S}_t,\mathcal{L}_t)$. To obtain a sample, a distribution or region-of-interest $\mathcal{Y}_t$ is sampled to obtain a transformation $\mathbf{y}^j_t \sim \mathcal{Y}_t$ that defines the state of the sample compared to the previous target location, $\mathbf{p}^j_{t-1} = \mathbf{p}_{t-1} \circ \mathbf{y}_t^j$, and the sample appearance is defined as $\mathbf{x}^j_t \in \mathcal{X}_t$. This sample is then evaluated by the classifier $\theta_t$ with scoring function $h: \mathcal{X}_t \rightarrow \mathbb{R}$,
\begin{equation}
s^j_t = h(\mathbf{x}_t^j | \theta_t).
\label{eq:score_single}
\end{equation}
Based on the score, a label $\ell^j_t$ is assigned to the sample. For supervised-learning classifiers \cite{avidan2004support}, the label is either positive (target) or negative (background), but semi-supervised classifiers (e.g., \cite{grabner2008semi,saffari2010robust}) or multi-instance learning (e.g., \cite{babenko2009visual,zhang2013robust}) allow the samples, which the classifier is uncertain about, to remain unlabeled by the classifier,
\begin{align}
\ell^j_t &=
  \begin{cases}   
   +1        & ,s^j_t > \tau_u \\
   -1        & ,s^j_t < \tau_l \\
   0         & ,\text{otherwise}
  \end{cases}
  \label{eq:label_single}
\end{align}
in which $\tau_l$ and $\tau_u$ denotes the lower and upper thresholds respectively. The unlabeled data are either discarded, used for later stages of tracking, or labeled by other mechanisms embedded in the tracker \cite{grabner2008semi,stalder2009beyond,tang2007co}. The target state, as mentioned, is estimated using $\psi (\mathcal{P}_t | \mathcal{S}_t,\mathcal{L}_t)$, and the classifier $\theta_t$ is updated by the all or a subset of the labeled data denoted by $\xi_t \subseteq \mathcal{P}_{\{1..t\}}$,
\begin{equation}
\theta_{t+1} = u(\theta_t, \mathcal{X}_{\xi_t}, \mathcal{L}_{\xi_t})
\end{equation}
where $u(.)$ is the model update function. The subset $\xi_t$ may involve all new data for online trackers ($\xi_t = \mathcal{P}_{t}$), a subset of the new data ($\xi_t \subset \mathcal{P}_{t}$) or recent data ($\xi_t \subset \mathcal{P}_{\{t-\Delta,...,t\}}$), and keyframe data ($\xi_t = \mathcal{P}_{\{k1,k2,...\}}$) \cite{meshgi2017active,hong2015multi}.

\begin{algorithm}[!t]
\DontPrintSemicolon
\SetKwInOut{Input}{input}\SetKwInOut{Output}{output}
\Input{Ensemble models $\mathcal{C} = \{\theta_{t}^{(c)}\}$}
\Input{Target position in previous frame $\mathbf{p}_{t-1}$}
\Output{Target position in current frame $\mathbf{p}_{t}$}
\BlankLine
\For{$j \leftarrow 1$ \KwTo $N_{samples}$ }
{
\emph{Draw sample $\mathbf{p}_t^j = \mathbf{p}_{t-1}^j \circ \mathbf{y}_t^j$ s.t. $\mathbf{y}_t^j \sim \mathcal{Y}_t$}\;
\emph{Calculate the classification score of members of $\mathcal{C}$}\;
\emph{Indicate the uncertainty flag $z_t^{(c,j)}$} (eq\eqref{eq:flag_ace})\;
\emph{Calculate ensemble's score $s^j_t$ and label $\ell^j_t$} (eq\eqref{eq:score_ace})\;
\emph{Calculate sample's informativeness $\eta^j_t$}\;
}
\For{$c \leftarrow 1$ \KwTo $n$ }
{
\emph{Obtain the error $e_t^{(c)}$ and weight $\alpha_t^{(c)}$} (eq\eqref{eq:error_ace}\eqref{eq:weight_ace})\;
\emph{Obtain the informative sample set $\mathcal{D}_t^{(c)}$}\;
\emph{Update the classifier $\theta_{t+1}^{(c)}$} (eq\eqref{eq:update_ace})\;
}
\If(target is not occluded){$\frac{1}{n}\sum_{c=1}^n e^{(c)}_t \leq \tau_{occ}$}
{
\emph{Estimate target state $\mathbf{\hat{p}}_t$} (eq\eqref{eq:state_ace}) \;
}

\BlankLine
\caption{ACET}
\label{alg:acet}
\vspace{-0.5cm}
\end{algorithm}

\subsection{Ensemble Discriminative Tracking}
\label{sec:ensemble}

A popular approach to robustify the classification in tracking-by-detection framework is to construct an ensemble of different (weak) classifiers $\mathcal{C}=\{\theta_t^{(1)},\ldots,\theta_t^{(n)}\}$, and combine their opinion about a sample by voting,
\begin{equation}
s^j_t = \sum_{c=1}^n \mathrm{sign} \big( h(\mathbf{x}_t^j | \theta_t^{(c)}) \big).
\label{eq:score_ensemble}
\end{equation}
In most of the cases, the weak classifiers are linearly combined with different associated weights,
\begin{equation}
s^j_t = \sum_{c=1}^n \alpha_t^{(c)} \mathrm{sign} \big( h(\mathbf{x}_t^j | \theta_t^{(c)}) \big),
\label{eq:score_ensemble_boost}
\end{equation}
where the weights $\alpha^{(c)}_t \in A_t$ are tuned using boosting \cite{avidan2007ensemble,grabner2006real,meshgi2016robust} or Bayesian treatment \cite{bai2013randomized}. A larger weight implies that the corresponding classifier of th eensemble is more discriminative, hence more useful. The labels are calculated from eq\eqref{eq:label_single} with $\tau^c_l$ and $\tau^c_u$ as the lower and upper thresholds for the ensemble score.

Finally, each classifier's model is updated independently,
\begin{equation}
\theta^{(c)}_{t+1} = u(\theta^{(c)}_t, \mathcal{X}_{\xi_t}, \mathcal{L}_{\xi_t})
\label{eq:update_ensemble}
\end{equation}
indicating that all of the ensemble members are trained with a similar set of samples $\xi_t$.

\subsection{Co-Training}
\label{sec:cotraining}
Built on co-training principle \cite{blum1998combining}, collaborative tracking (co-tracking) provides a framework in which the classifiers exchange their information to promote tracking results and break self-learning loop. In this two-classifier framework \cite{tang2007co}, the challenging samples for one classifier is labeled by the other one, i.e., if a classifier finds a sample difficult to label, it relies on the other classifier to label it for this frame and similar samples in the future.
\begin{align}
s^j_t &=
  \begin{cases}   
   \alpha^{(2)}_t h(\mathbf{x}_t^j | \theta_t^{(2)})        & ,\tau_l < h(\mathbf{x}_t^j | \theta_t^{(1)}) < \tau_u \\
   \alpha^{(1)}_t h(\mathbf{x}_t^j | \theta_t^{(1)})        & ,\tau_l < h(\mathbf{x}_t^j | \theta_t^{(2)}) < \tau_u \\
   \sum_{c=1}^2\alpha^{(c)}_t h(\mathbf{x}_t^j | \theta_t^{(c)})         & ,\text{otherwise}
  \end{cases}
  \label{eq:score_co}
\end{align}
The collaborative label is obtained by applying eq\eqref{eq:label_single} on this score. The weights of the classifiers are adjusted by comparing the labels of each classifier to the collaborative label. Eventually, the trackers which label a sample are getting updated by it. 

\subsection{Active Collaborative Ensemble Tracker}
\label{sec:acet}
The proposed tracker, ACET, is an ensemble tracker in which the co-training rule provides the samples for retraining each classifier, and active learning selects the most informative ones to improve the generalization and efficiency of the model update. Furthermore, by forgetting older samples with different memory horizons, the ensemble is diversified and non-stationary target appearance distributions are better accommodated.

Here, the ensemble $\mathcal{C}=\{\theta_t^{(1)},\ldots,\theta_t^{(n)}\}$ is constructed of similar classifiers but with different memory spans $\{ \Delta^{(1)},\ldots,\Delta^{(n)} \}$. Sample $\mathbf{x}^j_t$ is obtained from a Gaussian field centered on last target state, $\mathcal{Y}_t = \mathcal{N}(\mathbf{p}_{t-1},\Sigma_{search})$. This sample is then scored by all members of the ensemble. Those members that are uncertain about labeling the sample are marked by flag $z_t^{(c,j)} \in \mathcal{Z}^{(c)}_t$,
\begin{align}
z_t^{(c,j)} &=
  \begin{cases}   
   0        & ,\tau_l < h(\mathbf{x}_t^j | \theta_t^{(c)}) < \tau_u \\
   1        & ,\text{otherwise}
  \end{cases}
  \label{eq:flag_ace}
\end{align}
which in turn helps to calculate the score of ensemble,
\begin{equation}
s^j_t = \sum_{c=1}^n \alpha_t^{(c)} z_t^{(c,j)} \mathrm{sign} \big( h(\mathbf{x}_t^j | \theta_t^{(c)}) \big),
\label{eq:score_ace}
\end{equation}
and label it using eq\eqref{eq:label_single} with $\tau^c_l$ and $\tau^c_u$ as thresholds.

Since the number of samples are limited, an approximation of the target location $\mathbf{\hat{p}}_t$ is obtained by calculating the expectation of target, i.e., by taking a weighted average of the target candidates (i.e., positive samples).
\begin{align}
\mathbf{\hat{p}}_t = \;\mathbb{E} [ \mathbf{p}_t^j ] = \sum_{\forall j, \ell^j_t>0} s^j_t \mathbf{p}_t^j.
\label{eq:state_ace}
\end{align}

Following the rule of co-training, only the classifiers that engaged in labeling a sample ($z_t^{(c,j)}=1$) should be updated with that sample. However, not all the samples are equally useful to train the ensemble. For instance, a sample for which half of the ensemble are uncertain about its label would be better for training compared to a sample for which only one of the classifiers is uncertain. To measure the ``informativeness'' of a sample, we count the number of the classifiers that elicit a strong belief about its label, $\eta_t^j = \sum_{c=1}^n z_t^{(c,j)}$. Then for training of each classifier of the ensemble, based on query-by-committee concept \cite{seung1992query}, those samples with $z_t^{(c,j)}=1$ are sorted based on $\eta_t^j$ and the first $m$ are used for retraining (stored in $\mathcal{D}^{(c)}_t$).
\begin{equation}
\theta^{(c)}_{t+1} = u(\theta^{(c)}_t, \mathcal{D}^{(c)}_{\{t-\Delta^{(c)},...,t\}}, \mathcal{L'}^{(c)}_{\{t-\Delta^{(c)},...,t\}})
\label{eq:update_ace}
\end{equation}
where $\mathcal{L'}^{(c)}_t$ contains the labels of the samples in $\mathcal{D}^{(c)}_t$. As a result, the diversity of the ensemble is increased by co-training, selective updating, and different memory horizons.

The weights of the classifier is calculated based on its agreement with the whole ensemble. The error of each classifier is determined by
\begin{equation}
e^{(c)}_t = \sum_j \mathds{1}\Big(\mathrm{sign} \big( h(\mathbf{x}^j_t | \theta^{(c)}_t) \big) \neq \ell^j_t \Big)
\label{eq:error_ace}
\end{equation}
in which $\mathds{1}(.)$ is the indicator function. Then the weight of each classifier is calculated as,
\begin{equation}
\alpha_t^{(c)} = 1 - \frac{e^{(c)}_t + \epsilon}{\sum_{i=1}^n e^{(i)}_t + \epsilon}
\label{eq:weight_ace}
\end{equation}
in which $\epsilon$ is a small constant. If the error average of the ensemble is very high, $\frac{1}{n}\sum_{c=1}^n e^{(c)}_t > \tau_{occ}$, then the target is likely to be mostly occluded.
Algorithm \ref{alg:acet} and Figure \ref{fig:schematic} summarize the proposed tracker.

\begin{figure}[!t]   
\centering
\includegraphics[width= 1\linewidth]{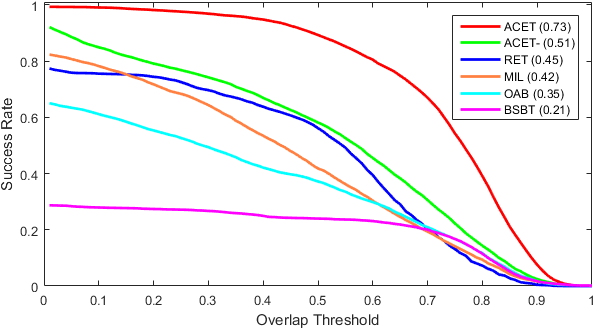}
\caption{Quantitative evaluation of ensemble trackers using success plot for 13 video sequence.}
\label{fig:ensemble}
\vspace{-0.5 cm}
\end{figure}

\section{Experiments}
\label{sec:bkg}

The proposed framework is comprised of several parameters: \textit{(i)} Sampling parameters such as number $N_{samples}$ and sampling distribution covariance $\Sigma_{search}$, \textit{(ii)} Ensemble parameters such as classifier count $n$, their memory spans $\Delta^{(c)}_t$ and labeling thresholds $\tau_l,\tau_u,\tau^{(c)}_l,\tau^{(c)}_u$, and \textit{(iii)} Tracking parameters such as occlusion threshold $\tau_{occ}$ and retraining subset size $m$.
On a P-IV PC at 3.5 GHz, ACET achieved 37.16 fps with a Mathlab/C++ implementation. 

\begin{figure}[!t]
\centering
\includegraphics[width= 0.32\linewidth]{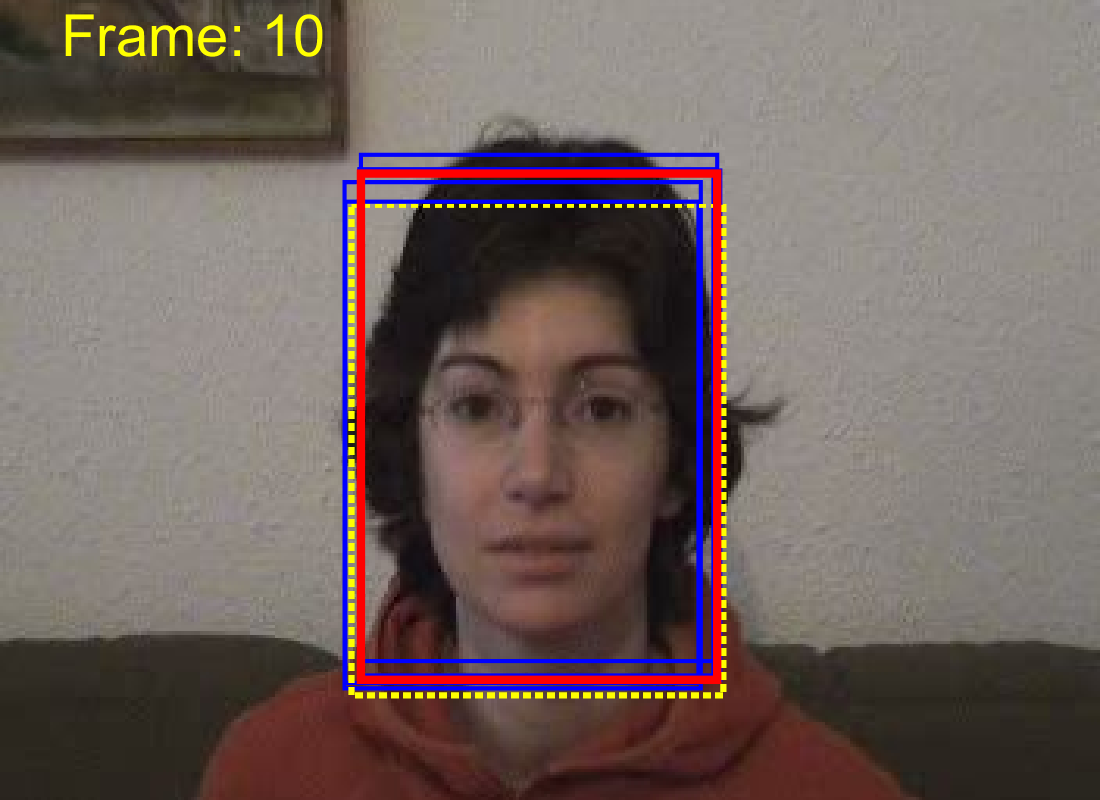}
\includegraphics[width= 0.32\linewidth]{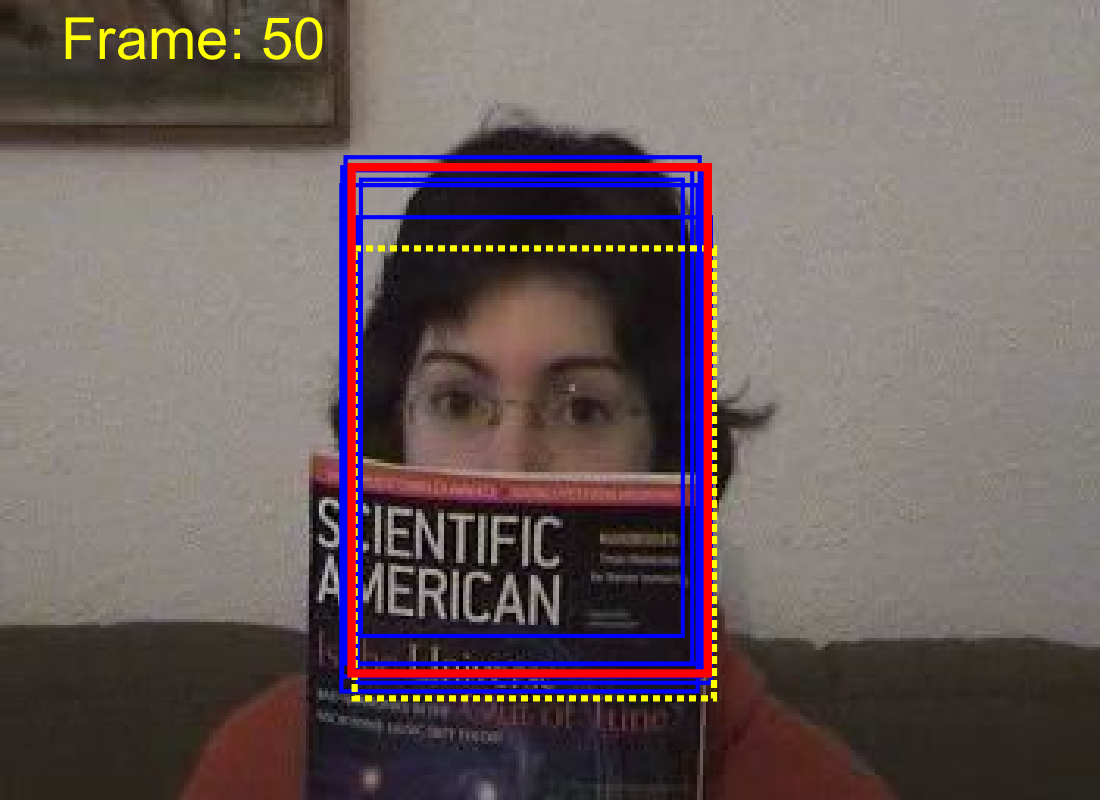}
\includegraphics[width= 0.32\linewidth]{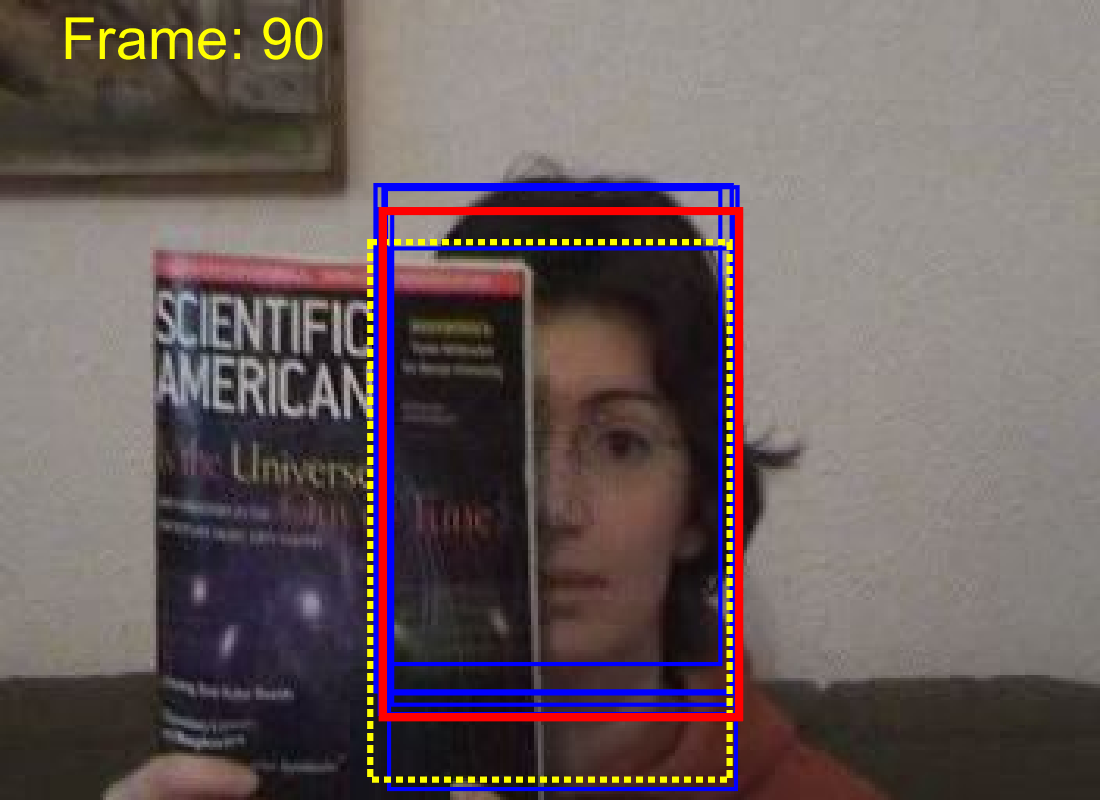}
\includegraphics[width= 0.32\linewidth]{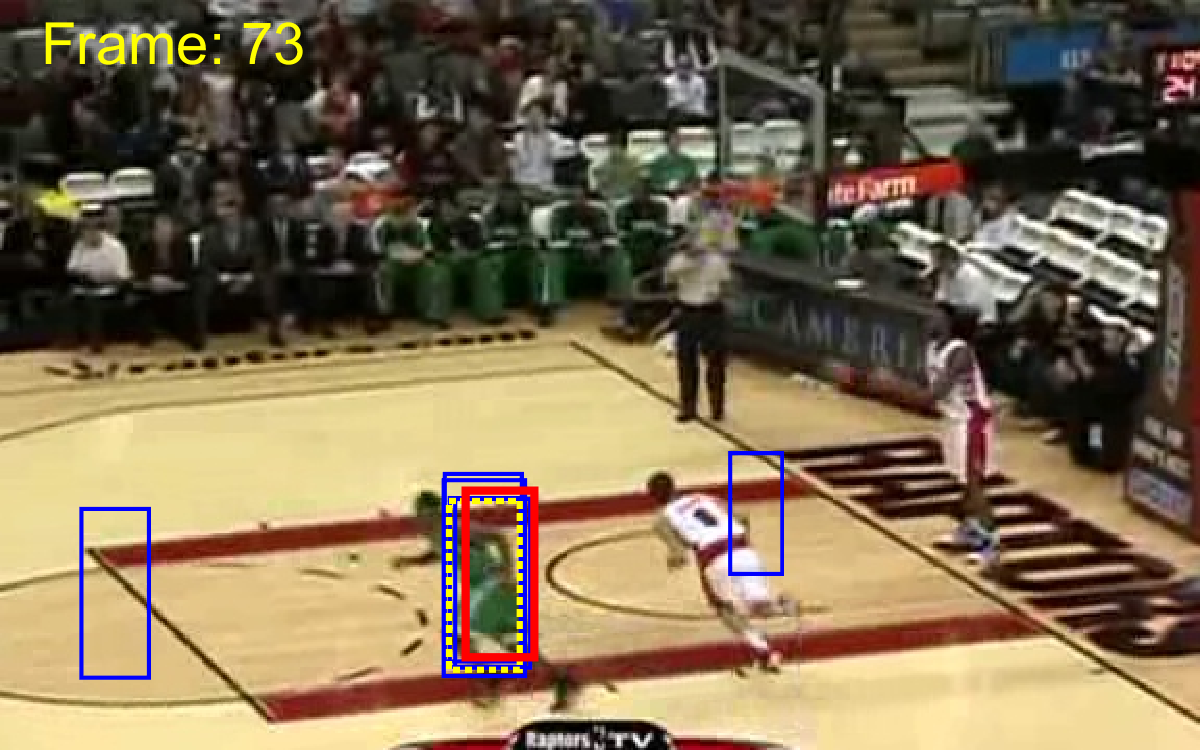}
\includegraphics[width= 0.32\linewidth]{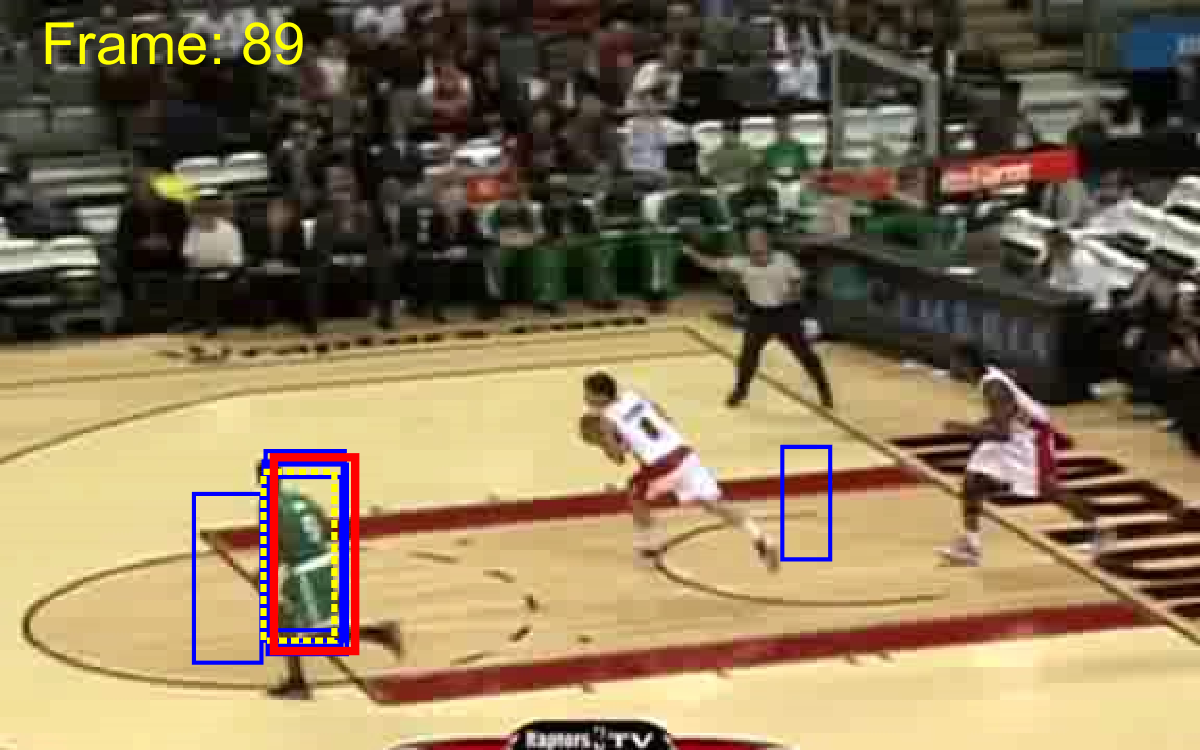}
\includegraphics[width= 0.32\linewidth]{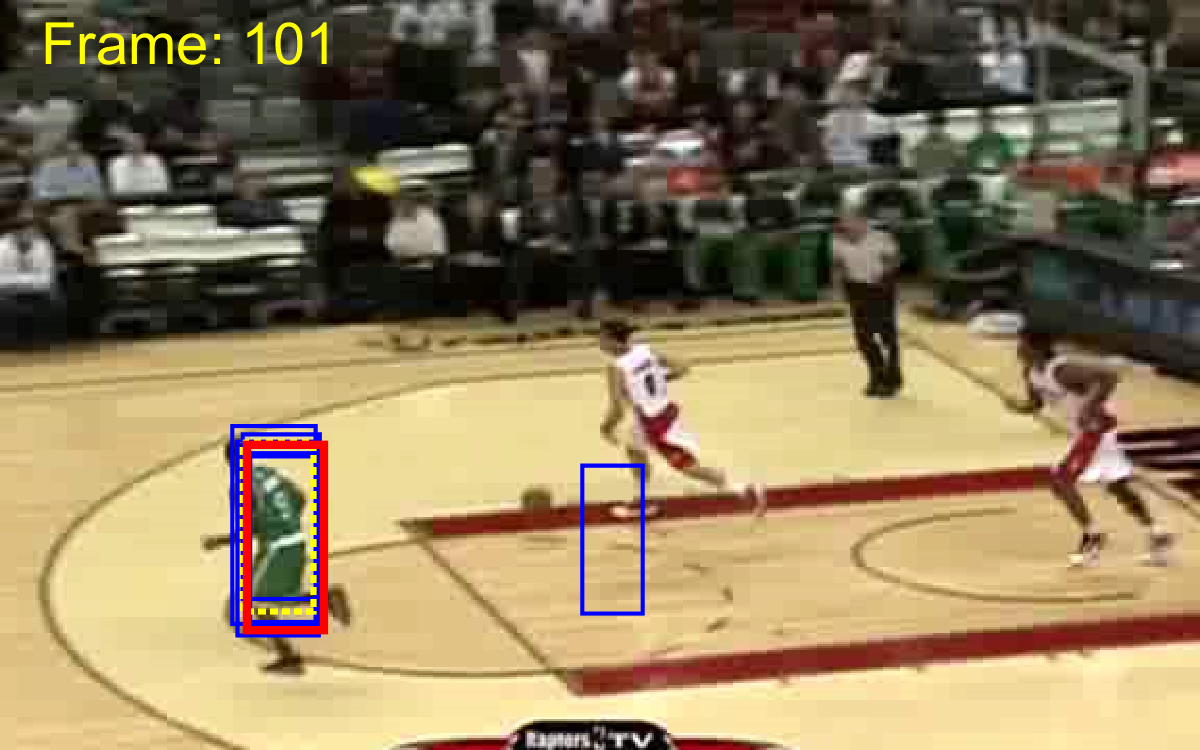}
\includegraphics[width= 0.32\linewidth]{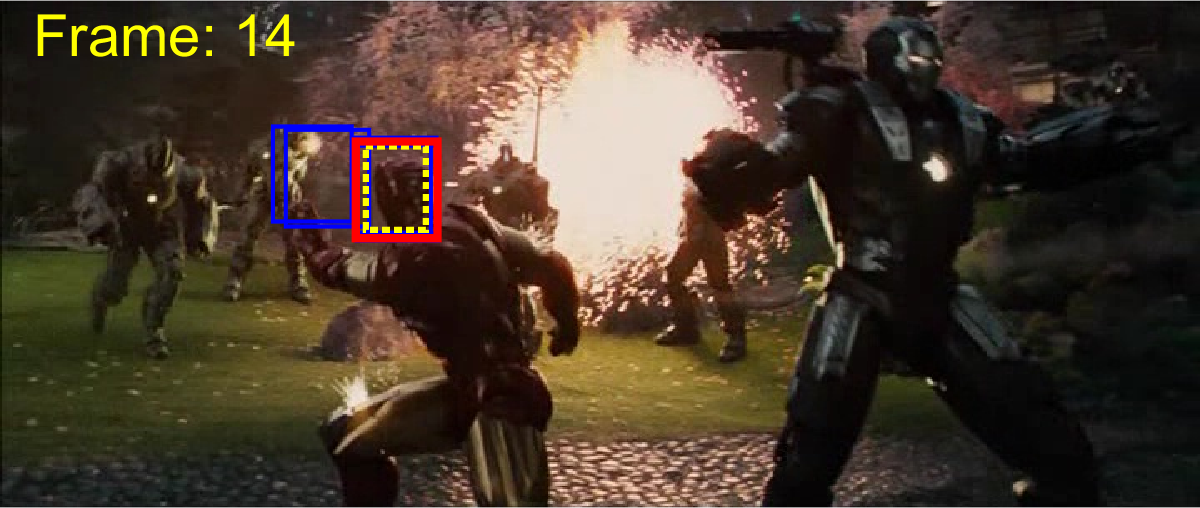}
\includegraphics[width= 0.32\linewidth]{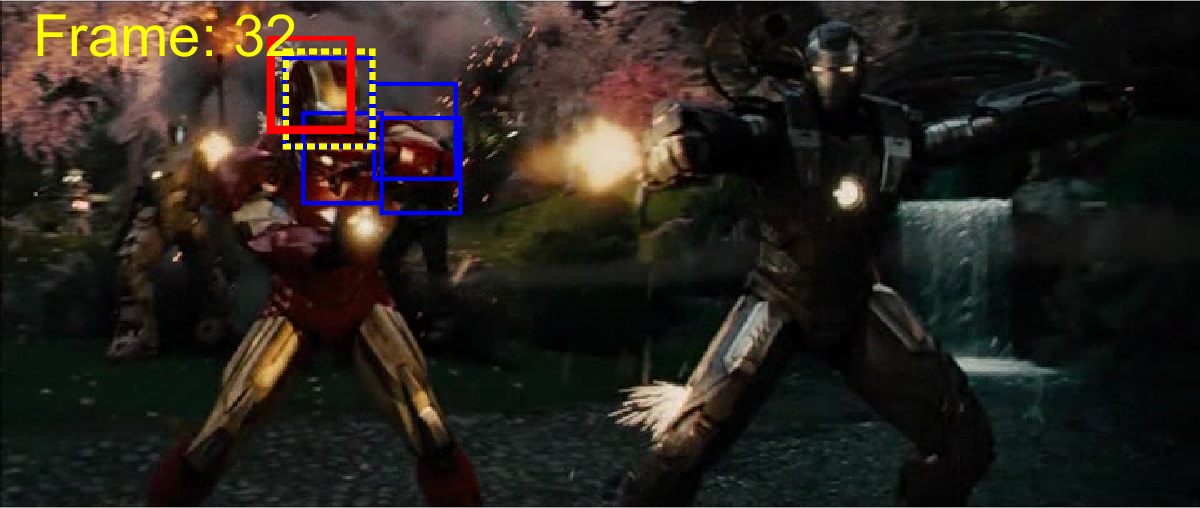}
\includegraphics[width= 0.32\linewidth]{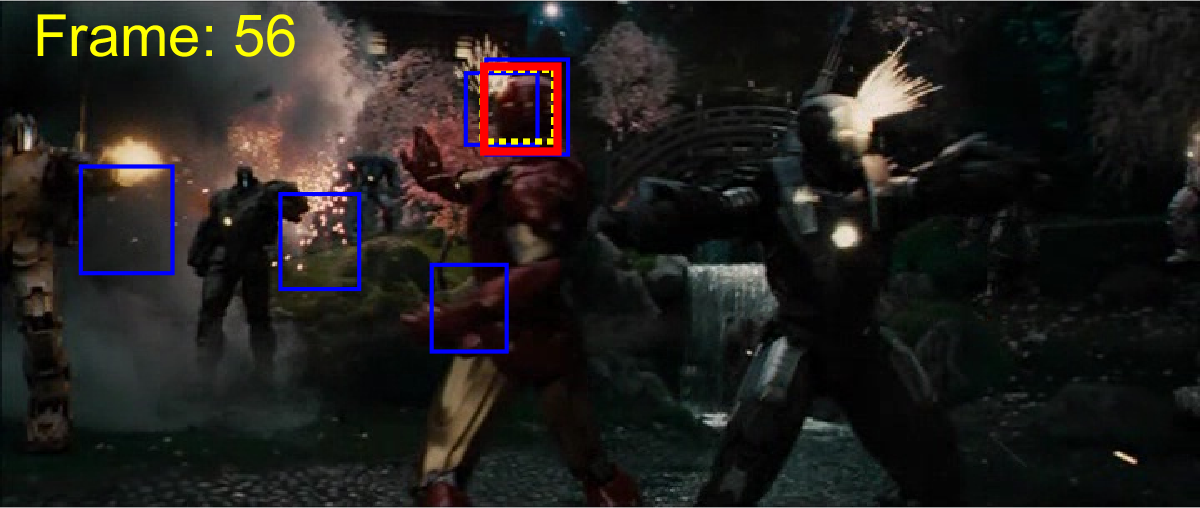}
\includegraphics[width= 0.32\linewidth]{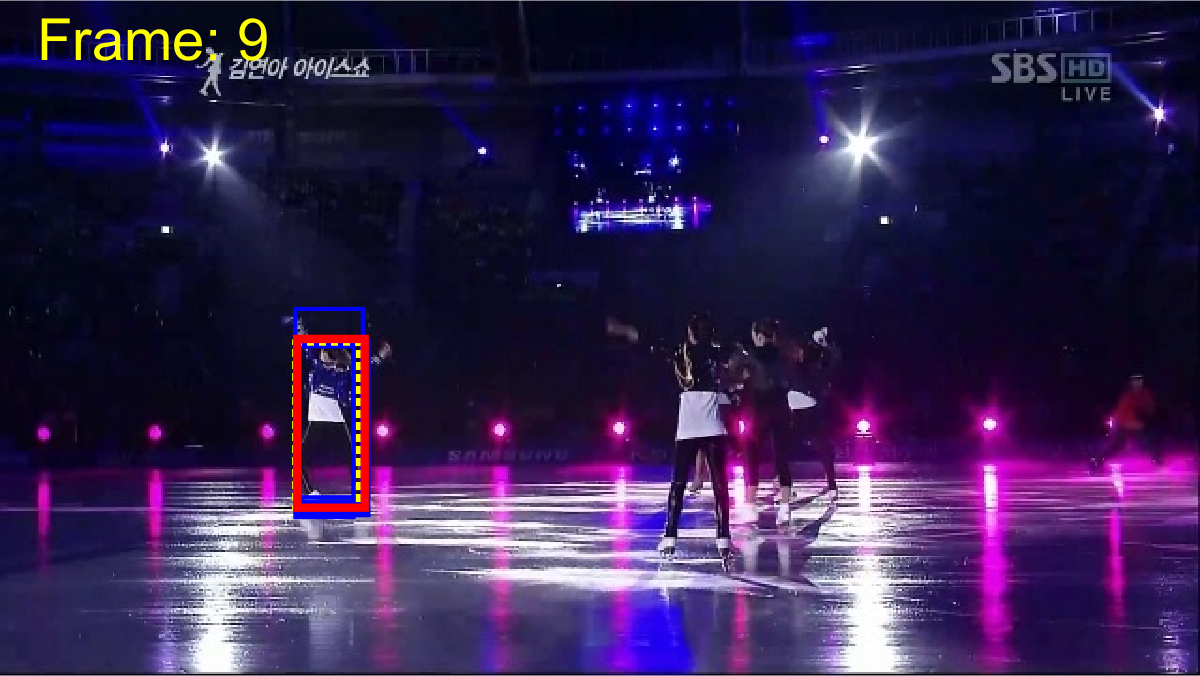}
\includegraphics[width= 0.32\linewidth]{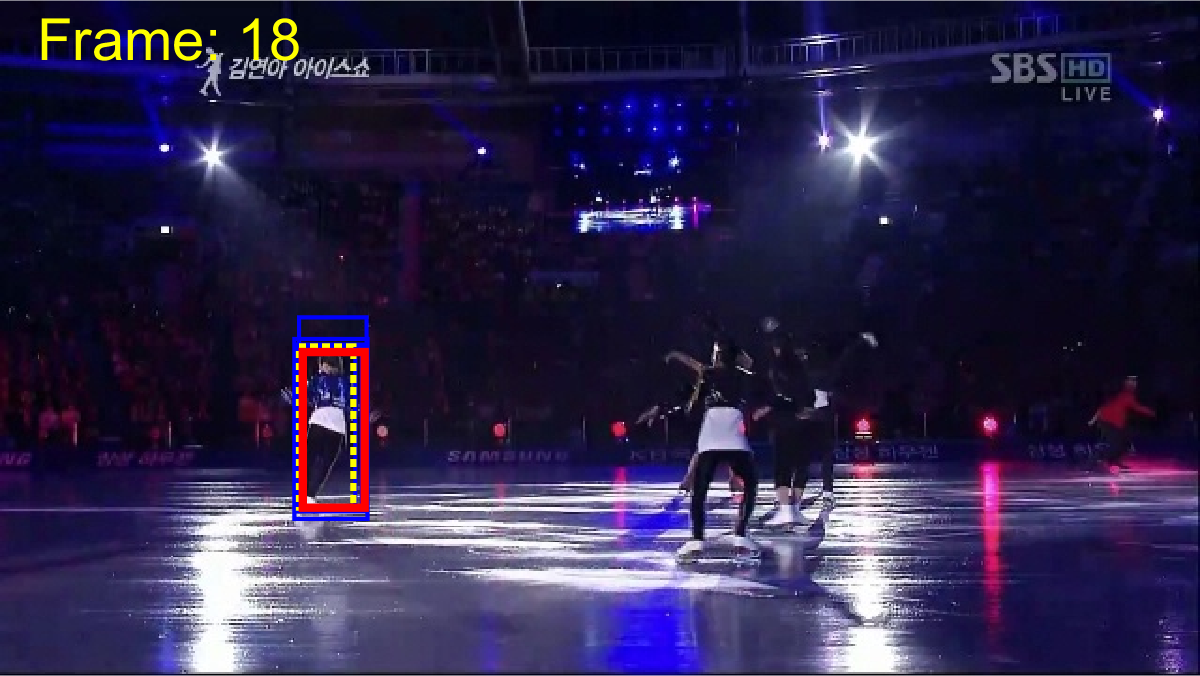}
\includegraphics[width= 0.32\linewidth]{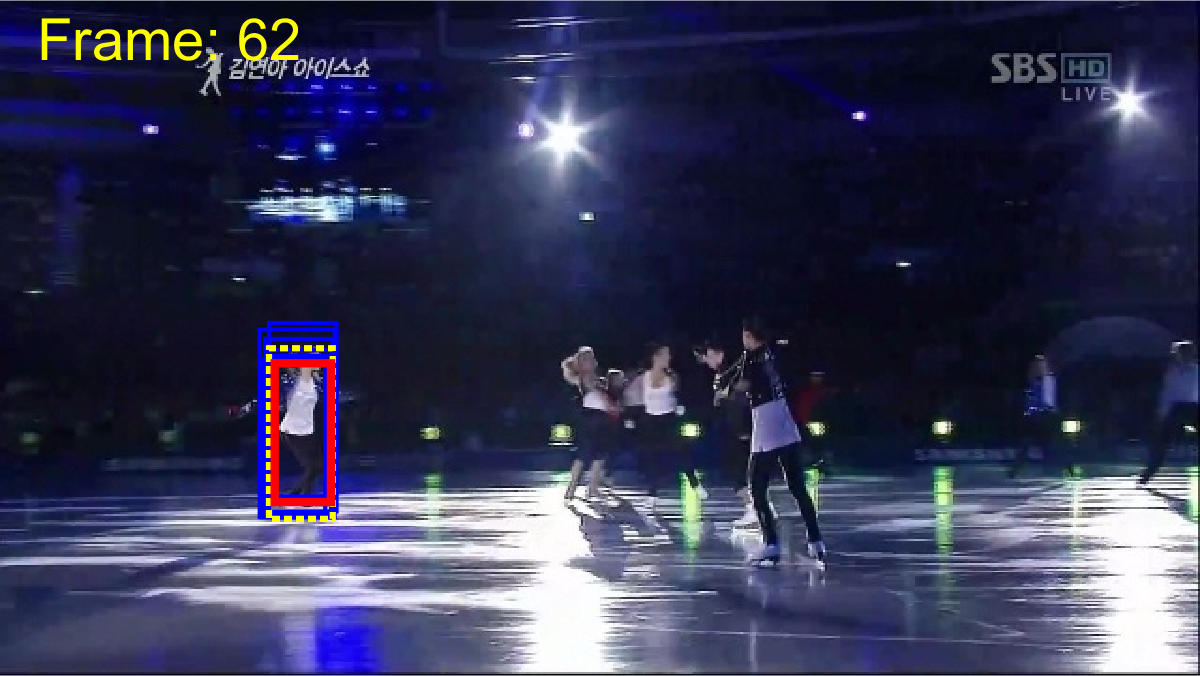}
\includegraphics[width= 0.32\linewidth]{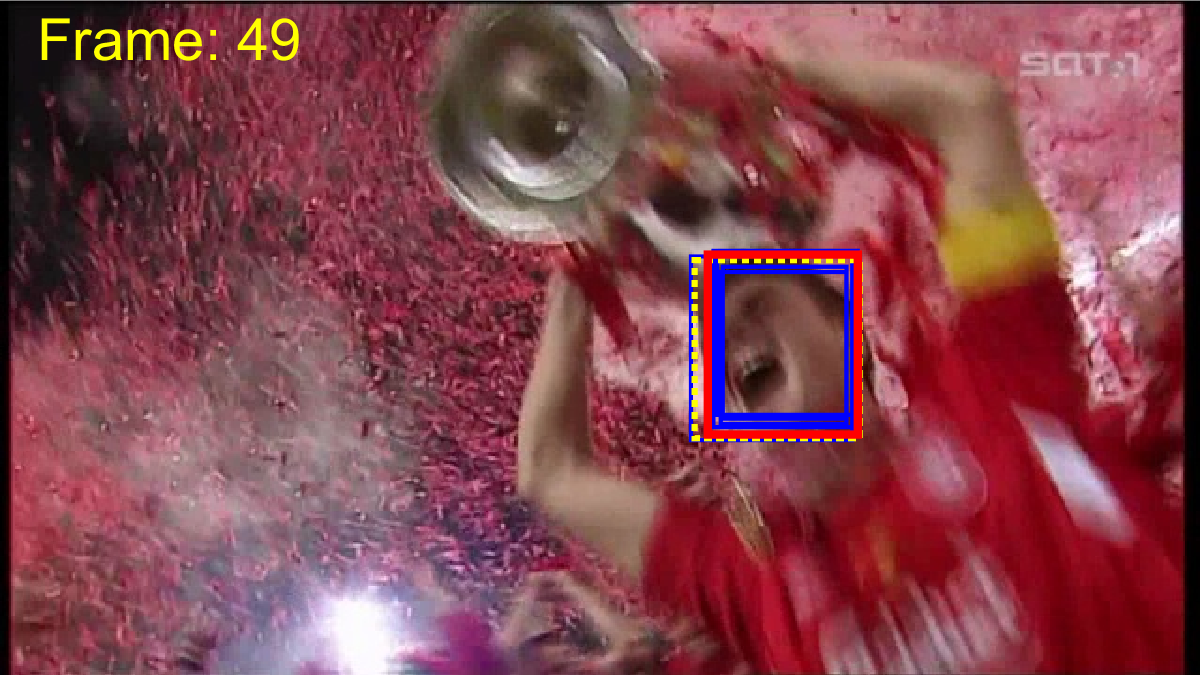}
\includegraphics[width= 0.32\linewidth]{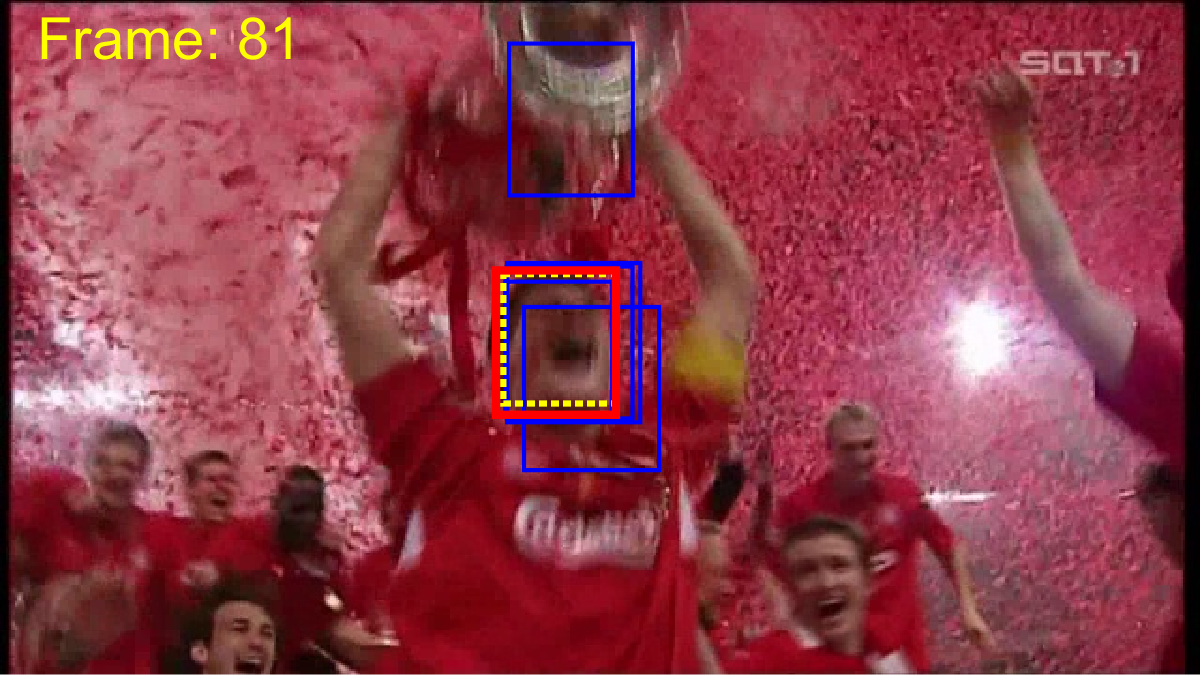}
\includegraphics[width= 0.32\linewidth]{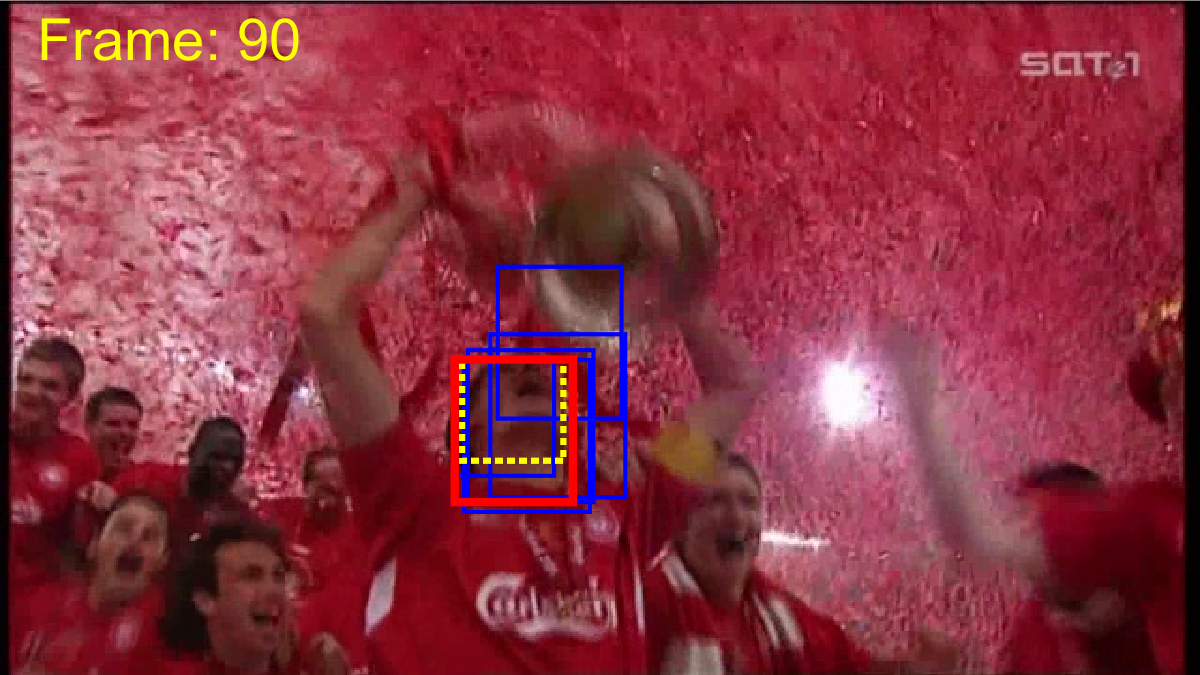}
\caption{Qualitative results of evaluated algorithms on several challenging videos \textbf{(top to down)} \textit{FaceOcc1} with severe occlusions, \textit{Basketball} with deformations, \textit{Ironman} with extreme rotations, \textit{Skating1} with drastic illumination changes, and \textit{Soccer} with background clutter. In these sequences the red box depicts the ACET against other trackers (blue). The ground truth is illustrated with yellow box. The results are available in the \texttt{webpage}.
}
\label{fig:eval_qual}
\vspace{-0.5 cm}
\end{figure}

\subsection{Evaluation Protocol}
\label{sec:protocol}
To evaluate the tracker, we employ success plot which measures the performance of a tracker which is a combination of its accuracy, reliability, and scale adaptation. 

The experiments are conducted to object tracking benchmark videos \cite{wu2013online}, which become a de-facto standard in comparing the performance of the trackers, and includes several subcategories, exploiting the performance of the trackers against various visual tracking challenges: illumination and scale variations (IV, SV), in- and out-of-plane rotations (IPR, OPR), fast motion and motion blur (FM, MB), deformations and low-resolution (DEF, LR), occlusion and shear problem (OCC, OV), and background clutter (BC).

\subsection{Comparison with other Ensemble Trackers}
For this experiment, we compare the proposed tracker (ACET) with online boosting tracker (OAB \cite{grabner2006real}) that utilizes different features to construct weak classifiers as ensemble members, and randomized ensemble tracker (RET \cite{bai2013randomized}) that make different strong classifiers out of a pool of weak classifiers, and construct the ensemble out of those strong classifiers. We also include MIL \cite{babenko2009visual} and BSBT\cite{stalder2009beyond} to represent ensemble trackers based on semi-supervised and multi-instance learning. Here, we implement a version of our tracker (ACET-) that use the same feature set to construct different members of the ensemble and the active learning and memory horizon subsampling is disabled. For the sake of compatibility with published RET results, 13 overlapping sequences with OTB-50 have been used.

Figure \ref{fig:ensemble} illustrate that the proposed framework works better than other ensemble methods regardless of the ensemble member construction. Yet, it is evident that using all features along with subsampling schemes for re-training classifier (by active learning and different memory spans) significantly improve the tracking performance.

\subsection{Comparison with State-of-the-art}
To provide a fair comparison, we compared ACET with state-of-the-art tracking-by-detection algorithms TLD \cite{kalal2012tracking}, STRK \cite{hare2011struck}, MEEM \cite{zhang2014meem}, correlation filter trackers SRDCF \cite{danelljan2015learning}, CCOT \cite{danelljan2016beyond} and multi-memory tracker MUSTer \cite{hong2015multi}. The comparison based on the area under curve of the success plot is presented in Table \ref{tab:attributes}. It is evident that ACET outperforms the other trackers in most of the categories and in total performance over the 50 videos. Yet, it should be noted that since the tracker utilized two features that are both sensitive to low resolution target appearances, (as expected) it is not able to perform well in LR category. The good performance of the tracker in target appearance change categories (IV, DEF, OCC, OV) can be attributed to the robustness of ensemble due to co-learning, while the good results on transformation categories (SV, IPR, OPR) can be attributed to good generalization obtained by active learning sample selection for ensemble retraining. Different memory spans helped the tracker to dominate motion categories (FM, MB), and a robust diverse ensemble obtained by all of these approaches resolved background clutter (BC) effectively. The quality of results are showed in Figure \ref{fig:eval_qual}.

\begin{table}[b]
\caption{Quantitative evaluation of state-of-the-art under different visual tracking challenges using AUC of success plot. The best performance for each attribute is shown in \textbf{bold}.}
\label{tab:attributes}
\centering
\scalebox{0.85}{
\renewcommand{\arraystretch}{1.1}
\begin{tabular}{c|c c c c c c c}

\makebox[9mm]{Attribute} & \makebox[8mm]{TLD} & \makebox[8mm]{STRK} & \makebox[8mm]{MEEM}& \makebox[8mm]{MUSTer}& \makebox[8mm]{SRDCF}& \makebox[8mm]{CCOT}& \makebox[8mm]{Ours} \\ \hline
IV     & 0.48 & 0.53 & 0.62 & 0.73 & 0.70 & 0.75 & \textbf{0.78} \\
DEF    & 0.38 & 0.51 & 0.62 & \textbf{0.69} & 0.67 & \textbf{0.69} & \textbf{0.69} \\
OCC    & 0.46 & 0.50 & 0.61 & 0.71 & 0.70 & 0.76 & \textbf{0.77} \\
SV     & 0.49 & 0.51 & 0.58 & 0.71 & 0.71 & 0.76 & \textbf{0.77} \\
IPR    & 0.50 & 0.54 & 0.58 & 0.69 & 0.70 & 0.72 & \textbf{0.77} \\
OPR    & 0.48 & 0.53 & 0.62 & 0.70 & 0.69 & 0.74 & \textbf{0.77} \\
OV     & 0.54 & 0.52 & 0.68 & 0.73 & 0.66 & 0.79 & \textbf{0.84} \\
FM     & 0.45 & 0.52 & 0.65 & 0.65 & 0.63 & 0.72 & \textbf{0.79} \\
MB     & 0.41 & 0.47 & 0.63 & 0.65 & 0.69 & 0.72 & \textbf{0.77} \\
BC     & 0.39 & 0.52 & 0.67 & 0.72 & 0.80 & 0.70 & \textbf{0.73} \\
LR     & 0.36 & 0.33 & 0.43 & 0.50 & 0.58 & \textbf{0.70} & 0.44 \\
\hline
ALL    & 0.49 & 0.55 & 0.62 & 0.72 & 0.70 & 0.75 & \textbf{0.76} \\

\end{tabular}
}
\vspace{-0.5cm}

\end{table}

\section{Conclusions}
\label{sec:conc}
In this study we proposed a novel framework for ensemble tracking, in which the classifiers exchange their data based on co-learning concept, and only the most informative samples are used for updating the classifiers to enhance generalization and accelerate convergence to non-stationary distributions of target appearance. Co-learning reduces the label noise, and break the self-learning loops that cause model drift, and together with different memory spans for the ensemble provides a robust model update scheme for ensemble tracking. The proposed tracker, ACET, outperformed other ensemble trackers and state-of-the-art on OTB-50\cite{wu2013online} database.


{\small
\bibliographystyle{ieee}
\bibliography{refs}

\begin{thebibliography}{10}\itemsep=-1pt

\bibitem{angelova2005pruning}
A.~Angelova, Y.~Abu-Mostafam, and P.~Perona.
\newblock Pruning training sets for learning of object categories.
\newblock In {\em CVPR'05}.

\bibitem{avidan2004support}
S.~Avidan.
\newblock Support vector tracking.
\newblock {\em PAMI}, 2004.

\bibitem{avidan2007ensemble}
S.~Avidan.
\newblock Ensemble tracking.
\newblock {\em PAMI}, 29, 2007.

\bibitem{babenko2009visual}
B.~Babenko, M.-H. Yang, and S.~Belongie.
\newblock Visual tracking with online multiple instance learning.
\newblock In {\em CVPR'09}, 2009.

\bibitem{bai2013randomized}
Q.~Bai, Z.~Wu, S.~Sclaroff, M.~Betke, and C.~Monnier.
\newblock Randomized ensemble tracking.
\newblock In {\em ICCV'13}, 2013.

\bibitem{bai2012robust}
Y.~Bai and M.~Tang.
\newblock Robust tracking via weakly supervised ranking svm.
\newblock In {\em CVPR'12}, 2012.

\bibitem{bao2012real}
C.~Bao, Y.~Wu, H.~Ling, and H.~Ji.
\newblock Real time robust l1 tracker using accelerated proximal gradient
  approach.
\newblock In {\em CVPR'12}.

\bibitem{bengio2009curriculum}
Y.~Bengio, J.~Louradour, R.~Collobert, and J.~Weston.
\newblock Curriculum learning.
\newblock In {\em ICML'09}, 2009.

\bibitem{blum1998combining}
A.~Blum and T.~Mitchell.
\newblock Combining labeled and unlabeled data with co-training.
\newblock In {\em COLT'98}, 1998.

\bibitem{danelljan2015learning}
M.~Danelljan, G.~Hager, F.~Shahbaz~Khan, and M.~Felsberg.
\newblock Learning spatially regularized correlation filters for visual
  tracking.
\newblock In {\em ICCV'15}, pages 4310--4318, 2015.

\bibitem{danelljan2016beyond}
M.~Danelljan, A.~Robinson, F.~S. Khan, and M.~Felsberg.
\newblock Beyond correlation filters: Learning continuous convolution operators
  for visual tracking.
\newblock In {\em ECCV'16}.

\bibitem{de2001robust}
F.~De~la Torre and M.~J. Black.
\newblock Robust principal component analysis for computer vision.
\newblock In {\em ICCV'01}, 2001.

\bibitem{fang2017online}
J.~Fang, H.~Xu, Q.~Wang, and T.~Wu.
\newblock Online hash tracking with spatio-temporal saliency auxiliary.
\newblock {\em CVIU}, 2017.

\bibitem{felzenszwalb2010object}
P.~F. Felzenszwalb, R.~B. Girshick, D.~McAllester, and D.~Ramanan.
\newblock Object detection with discriminatively trained part-based models.
\newblock {\em PAMI}, 32(9):1627--1645, 2010.

\bibitem{gall2011hough}
J.~Gall, A.~Yao, N.~Razavi, L.~Van~Gool, and V.~Lempitsky.
\newblock Hough forests for object detection, tracking, and action recognition.
\newblock {\em PAMI}, 2011.

\bibitem{grabner2006real}
H.~Grabner, M.~Grabner, and H.~Bischof.
\newblock Real-time tracking via on-line boosting.
\newblock In {\em BMVC'06}, 2006.

\bibitem{grabner2008semi}
H.~Grabner, C.~Leistner, and H.~Bischof.
\newblock Semi-supervised on-line boosting for robust tracking.
\newblock In {\em ECCV'08}. 2008.

\bibitem{grabner2010tracking}
H.~Grabner, J.~Matas, L.~Van~Gool, and P.~Cattin.
\newblock Tracking the invisible: Learning where the object might be.
\newblock In {\em CVPR'10}, 2010.

\bibitem{hare2011struck}
S.~Hare, A.~Saffari, and P.~H. Torr.
\newblock Struck: Structured output tracking with kernels.
\newblock In {\em ICCV'11}, 2011.

\bibitem{henriques2012exploiting}
J.~F. Henriques, R.~Caseiro, P.~Martins, and J.~Batista.
\newblock Exploiting the circulant structure of tracking-by-detection with
  kernels.
\newblock In {\em ECCV'12}, pages 702--715. Springer, 2012.

\bibitem{hong2015multi}
Z.~Hong, Z.~Chen, C.~Wang, X.~Mei, D.~Prokhorov, and D.~Tao.
\newblock Multi-store tracker (muster): a cognitive psychology inspired
  approach to object tracking.
\newblock In {\em CVPR'15}.

\bibitem{kalal2012tracking}
Z.~Kalal, K.~Mikolajczyk, and J.~Matas.
\newblock Tracking-learning-detection.
\newblock {\em PAMI}, 34(7):1409--1422, 2012.

\bibitem{kiani2017learning}
H.~Kiani~Galoogahi, A.~Fagg, and S.~Lucey.
\newblock Learning background-aware correlation filters for visual tracking.
\newblock {\em arXiv}, 2017.

\bibitem{kiani2015correlation}
H.~Kiani~Galoogahi, T.~Sim, and S.~Lucey.
\newblock Correlation filters with limited boundaries.
\newblock In {\em CVPR'15}, 2015.

\bibitem{kwon2017leveraging}
J.~Kwon, R.~Timofte, and L.~Van~Gool.
\newblock Leveraging observation uncertainty for robust visual tracking.
\newblock {\em CVIU}, 2017.

\bibitem{lapedriza2013all}
A.~Lapedriza, H.~Pirsiavash, Z.~Bylinskii, and A.~Torralba.
\newblock Are all training examples equally valuable?
\newblock {\em arXiv}, 2013.

\bibitem{leistner2010miforests}
C.~Leistner, A.~Saffari, and H.~Bischof.
\newblock Miforests: Multiple-instance learning with randomized trees.
\newblock In {\em ECCV'10}, 2010.

\bibitem{leistner2009robustness}
C.~Leistner, A.~Saffari, P.~Roth, and H.~Bischof.
\newblock On robustness of on-line boosting: a competitive study.
\newblock In {\em ICCVw'09}.

\bibitem{matthews2004template}
I.~Matthews, T.~Ishikawa, and S.~Baker.
\newblock The template update problem.
\newblock {\em PAMI}, 2004.

\bibitem{meshgi2016data}
K.~Meshgi, S.-I. Maeda, S.~Oba, and S.~Ishii.
\newblock Data-driven probabilistic occlusion mask to promote visual tracking.
\newblock In {\em CRV'16}.

\bibitem{meshgi2017active}
K.~Meshgi, S.~Oba, and S.~Ishii.
\newblock Active discriminative tracking using collective memory.
\newblock In {\em MVA'17}.

\bibitem{meshgi2016robust}
K.~Meshgi, S.~Oba, and S.~Ishii.
\newblock Robust discriminative tracking via query-by-committee.
\newblock In {\em AVSS'16}, 2016.

\bibitem{nam2016learning}
H.~Nam and B.~Han.
\newblock Learning multi-domain convolutional neural networks for visual
  tracking.
\newblock In {\em CVPR'16}.

\bibitem{oza2005online}
N.~C. Oza.
\newblock Online bagging and boosting.
\newblock In {\em SMC'05}, 2005.

\bibitem{perez2002color}
P.~P{\'e}rez, C.~Hue, J.~Vermaak, and M.~Gangnet.
\newblock Color-based probabilistic tracking.
\newblock In {\em ECCV'02}.

\bibitem{razavi2012latent}
N.~Razavi, J.~Gall, P.~Kohli, and L.~Van~Gool.
\newblock Latent hough transform for object detection.
\newblock {\em ECCV'12}.

\bibitem{ross2008incremental}
D.~A. Ross, J.~Lim, R.-S. Lin, and M.-H. Yang.
\newblock Incremental learning for robust visual tracking.
\newblock {\em IJCV}, 2008.

\bibitem{saffari2010robust}
A.~Saffari, C.~Leistner, M.~Godec, and H.~Bischof.
\newblock Robust multi-view boosting with priors.
\newblock In {\em ECCV'10}. 2010.

\bibitem{saffari2009line}
A.~Saffari, C.~Leistner, J.~Santner, M.~Godec, and H.~Bischof.
\newblock On-line random forests.
\newblock In {\em ICCVw'09}.

\bibitem{santner2010prost}
J.~Santner, C.~Leistner, A.~Saffari, T.~Pock, and H.~Bischof.
\newblock Prost: Parallel robust online simple tracking.
\newblock In {\em CVPR'10}.

\bibitem{seung1992query}
H.~S. Seung, M.~Opper, and H.~Sompolinsky.
\newblock Query by committee.
\newblock In {\em COLT'92}, pages 287--294. ACM, 1992.

\bibitem{stalder2009beyond}
S.~Stalder, H.~Grabner, and L.~Van~Gool.
\newblock Beyond semi-supervised tracking: Tracking should be as simple as
  detection, but not simpler than recognition.
\newblock In {\em ICCVw'09}.

\bibitem{taalimi2015online}
A.~Taalimi, H.~Qi, and R.~Khorsandi.
\newblock Online multi-modal task-driven dictionary learning and robust joint
  sparse representation for visual tracking.
\newblock In {\em AVSS'15}, 2015.

\bibitem{tang2007co}
F.~Tang, S.~Brennan, Q.~Zhao, and H.~Tao.
\newblock Co-tracking using semi-supervised support vector machines.
\newblock In {\em ICCV'07}.

\bibitem{vijayanarasimhan2011cost}
S.~Vijayanarasimhan and K.~Grauman.
\newblock Cost-sensitive active visual category learning.
\newblock {\em IJCV}, 2011.

\bibitem{wu2013online}
Y.~Wu, J.~Lim, and M.-H. Yang.
\newblock Online object tracking: A benchmark.
\newblock In {\em CVPR'13}, pages 2411--2418. IEEE, 2013.

\bibitem{wu2015robust}
Y.~Wu, M.~Pei, M.~Yang, and Y.~Jia.
\newblock Robust discriminative tracking via landmark-based label propagation.
\newblock {\em TIP}, 2015.

\bibitem{zhang2014meem}
J.~Zhang, S.~Ma, and S.~Sclaroff.
\newblock Meem: Robust tracking via multiple experts using entropy
  minimization.
\newblock In {\em ECCV'14}.

\bibitem{zhang2013real}
K.~Zhang and H.~Song.
\newblock Real-time visual tracking via online weighted multiple instance
  learning.
\newblock {\em PR}, 2013.

\bibitem{zhang2013robust}
K.~Zhang, L.~Zhang, M.-H. Yang, and Q.~Hu.
\newblock Robust object tracking via active feature selection.
\newblock {\em CSVT}, 2013.

\bibitem{zhu2012we}
X.~Zhu, C.~Vondrick, D.~Ramanan, and C.~C. Fowlkes.
\newblock Do we need more training data or better models for object detection?.
\newblock In {\em BMVC'12}.

\end{thebibliography}
}


\end{document}